\documentclass{sig-alternate}
\usepackage{url}
\usepackage{multirow}

\newcommand{\argmin}{\operatornamewithlimits{argmin}}
\usepackage{times}
\usepackage{latexsym}
\usepackage{amsmath}
\usepackage{multirow}
\usepackage{graphicx}
\usepackage{multirow}
\usepackage{subfigure}
\usepackage{url}
\usepackage{comment}

\usepackage{color}
\usepackage{url}
\usepackage{comment}
\usepackage{enumitem}
  {\begin{itemize}[topsep=0pt, partopsep=0pt] \footnotesize%
    \setlength{\itemsep}{0pt}%
    \setlength{\parskip}{0pt}%
    }%
  {\end{itemize}}

  {\begin{itemize}[topsep=10pt, partopsep=0pt] \footnotesize%
    \setlength{\itemsep}{2pt}%
    \setlength{\parskip}{0pt}%
    }%
  {\end{itemize}}

  \usepackage{CJKutf8}

\title{Modeling Word Relatedness in Latent Dirichlet Allocation}
\begin{document}
\date{}

\numberofauthors{3} 
\author{
\alignauthor
Xun Wang\\
\affaddr{ Linguistic Intelligence Research Group. }\\
       \affaddr{NTT Communication Science Laboratories. }\\
       \affaddr{xunwang45@gmail.com}
}
\maketitle

\maketitle
\begin{abstract}
Standard LDA model suffers the problem that the topic assignment of each word is independent and word correlation hence is neglected.
To address this problem, in this paper, we propose a model called Word Related Latent Dirichlet Allocation (WR-LDA) by incorporating word correlation into LDA topic models. This leads to new capabilities that standard LDA model does not have such as estimating infrequently occurring words or multi-language topic modeling. Experimental results demonstrate the effectiveness of our model compared with standard LDA.
\end{abstract}
\begin{CJK*}{UTF8}{gbsn}

\section{Introduction}
Latent Dirichlet Allocation(LDA) (Blei et al., 2003) has been widely used in many different fields for its ability in capturing latent semantics in textual and image data. In standard LDA, the generating probability is estimated from a document-word matrix using word frequency information.
One main disadvantage of LDA is that topic assignment of words is conditionally independent from each other and the relevance between vocabularies (i.e semantic similarity) is totally neglected. This inability makes LDA fall short in many aspects. For example, it is quite plausible that synonymous words such as "use" and "utilize", "Chinese" and "China" or "politics" and "politicians" should be generated from the same topic. However, standard LDA models treat them as independent units and if they do not co-appear in the same context, they can hardly be included in the same topic. An extreme case is cross-lingual topic modeling where words in different languages never co-occur with each other, though talking about same topics as they are.

Recently, a number of attempts have been made to address the limitation due to conditional-independence. In cross-lingual topic extraction, researchers try to match word pairs and discover the correspondence aligned either at sentence level or document level \cite{zhao2006bitam,tam2007bilingual,mimno2009polylingual,ni2009mining}. Zhao and Xing \cite{zhao2006bitam} incorporated $word-pair$, $sentence-pair$ and $document-pair$ knowledge information into statistical models. One disadvantage is that these approaches usually require aligned text corpus based on machine translation techniques.
Boyd-Graber and Blei \cite{boyd2009multilingual} developed the unaligned topic approach called {\em MUTO} where topics are distributions over word pairs instead of just being distributions
over terms.
However, these methods focus on cross-lingual techniques where word correlation is only partly considered.

To the best of our knowledge, the closest existing works are the approach developed by Andrzejewski et al. \cite{andrzejewski2009incorporating} and the approach developed by Petterson et al. \cite{petterson2010word} where word relations are considered. In Andrzejewski et al.'s work \cite{andrzejewski2009incorporating}, domain knowledge is used to model vocabulary relations such as {\em Must-Link} or {\em Cannot-Link}. By applying a novel Dirichlet Tree prior, {\em Must-Link} words are more likely to be generated from the same topic and {\em Cannot-Link} are less likely. However, Andrzejewski et al.'s approach requires specific domain knowledge and can not be extended to general word correlation.
In Petterson et al. 's work\cite{petterson2010word}, they use a sophisticated biased prior in Dirichlet distribution to consider the word correlations rather than the uniform one used in Standard LDA. One disadvantage is that their work considers word correlations only in the prior, but not in the algorithm. 

In this paper, we propose an approach called Word Related Latent Dirichlet Allocation (WR-LDA) that incorporates word correlation into topic model. Given a collection of documents, standard LDA topic model wishes finding parameters by maximizing the marginal log likelihood of data. In our model, we sacrifice part of the maximization of log likelihood by incorporating vocabulary correlations based on the assumption that similar words tend to be generated from similar topics. We experiment our model on the different datasets and results indicate the effectiveness of model.

Section 2 presents some related work and Section 3 presents our model. The remaining is the experiments and the conclusion.

\section{Related Work}
Topic models such as Latent Dirichlet Allocation (LDA) \cite{blei2003latent} and PLSA (Probabilistic Latent Semantic Analysis) \cite{hofmann1999probabilistic} have been widely used in many different fields such as social analysis \cite{li2014timeline,li2014major,li2014weakly}, event detection \cite{aggarwal2012mining,niebles2008unsupervised}, text classification \cite{mcauliffe2008supervised,ramage2009labeled} or facet mining \cite{li2013identifying,titov2008modeling}.
for their ability in discovering topics latent in the document collection. Standard topic models suffer the disadvantage that researchers only use word co-appearance frequency for topic modeling without considering word correlations. However similar or synonymous two terms are, if they do not co-appear in the document, they can hardly be classified into the same topic. This disadvantage is largely magnified in multi-lingual topic modeling where different languages never co-occur with each other.

Previous work on multilingual topic models mostly require parallelism at either the sentence level or document level \cite{andrzejewski2009incorporating} and the approach developed by Petterson et al. \cite{petterson2010word}. Boyd-Graber and Blei (2009) proposed a multilingual topic approach which requires a list of word pairs. Topic is defined as distribution over word pairs. However, word correlation is only partly considered in multilingual topic techniques.

 Andrzejewski et al. \cite{andrzejewski2009incorporating}  proposed the approach by considering domain knowledge to model vocabulary relations such as {\em Must-Link} or {\em Cannot-Link}
by using a Dirichlet Tree prior in topic models. However, their model can not be extended to general word correlation.
 Petterson et al. \cite{petterson2010word}uses a biased Dirichlet prior by using a Logistic word smoother which takes accounts word relations.
 The disadvantage of their model is that their work take account of word correlations only in the prior, but not in the algorithm.

\section{LDA topic model}
\begin{figure}
\centering
\includegraphics[width=2.6in]{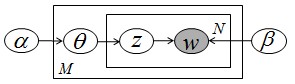}
\caption{Graphical model for LDA}
\end{figure}
The Latent Dirichlet allocation (LDA) model (Blei et al., 2003) defines the document likelihood
using a hierarchical Bayesian scheme. Specifically, each document is presented as a mixture of topics which is drawn from a Dirichlet distribution
and each topic is presented as a mixture of words.
Document words are sampled from a topic-specific
word distribution specified by a drawn of
the word-topic-assignment from the topic proportion
vector. Let
K be the number of topics; V be the number of the
terms in a vocabulary. $\beta$ is a $K\times V$ matrix and $\beta_k$ is the distribution vector over $V$ terms where $\beta_{ij}=P(w=j|z_w=i)$ denotes the probability word $w$ is generated from topic k.
The generation for LDA is shown as follows:

\noindent 1. For each document $m\in [1,M]$:\\
 Draw topic proportion $\theta_m|\alpha\sim Dir(\alpha)$\\
2. For each word $w$\\
~2.1 draw topic $z_{w}\sim Multinomial(\theta_m)$\\
~2.2 draw $w\sim p(w|z_{w},\beta)$\\

LDA can be inferred from a collapsed Gibbes sampling (Heinrich 2005) or variational inference (Blei et al., 2003). Variational inference tries to find parameter $\alpha$ and $\beta$ that maximize the (marginal) log likelihood
of the data:
\begin{equation}
L(\alpha,\beta)=\sum_{d=1}^M \log(w_d|\alpha,\beta)
\end{equation}
Since the posterior distribution of latent variables can not be computed efficiently, in variational inference of LDA, researchers use variational distribution $q(\theta,z|\gamma,\phi)$ to approximate posterior distribution in each document.
\begin{equation}
q(\theta,z|\gamma,\phi)=q(\theta_m|\gamma_m)\prod_nq(z_n|\phi_n)
\end{equation}
$\phi_n=\{\phi_{n1},\phi_{n2},...,\phi_{nK}\}$. $\phi_{nk}$ can be interpreted as the probability that word at $n^{th}$ position in current document is generated from topic $k$. The next step is to set up an optimizing problem to determine the value of $\gamma$ and $\phi$ where the desideratum of finding a tight lower bound on the log likelihood is transformed to the following optimization problem:
\begin{equation}
(\gamma,\phi)=\argmin_{\gamma,\phi}KL(q(\theta,z|\gamma,\phi)||p(\theta,z|w,\alpha,\beta))
\end{equation}
where $KL(p||q)$ denotes the  KullbackLeibler (KL) divergence between two distributions. 
\begin{equation}
\begin{aligned}
&L(\gamma,\phi;\alpha,\beta)=\\
&E_q[\log p(w|z,\beta)]+E_q[\log p(z|\theta)]+E_q[\log p(\theta|\alpha)]\\
&-E_q[\log q(\theta)]-E_q[\log q(z)]-E_q[\log q(\beta)]
\end{aligned}
\end{equation}
$\{\alpha,\beta\}$ can be iteratively inferred from variational EM algorithm (Blei et al., 2003).

\section{WR-LDA}
\subsection{Description}
In this section, we present our WR-LDA model and show how it can incorporate word correlation into topic models. Let $G=(V,E)$ denote the graph where $V=\{w_i\}_{i=1}^{i=V}$ denotes word collection and $E=\{e_{w,w'}:w\in V, w'\in V\}$. $e(w,w')$ denotes the edge between node $w$ and $w'$. $\kappa_{w,w'}$ denotes the edge weight which is the similarity between word $w$ and $w'$. Based on the assumption that similar words should have similar probability generated by different topics, we introduce the $Loss~function~R(\beta)$, inspired by the idea of graph harmonic function (Zhu et al., 2003; Mei et al., 2008).
\begin{equation}
R(\beta)=\frac{1}{2}\sum_w\sum_{w'}\sum_k\kappa_{w,w'}(\beta_{kw}-\beta_{kw'})^2
\end{equation}
Intuitively, $R(\beta)$ measures the difference between $p(w|\beta)$ and $p(w'|\beta)$ for each pair $(w,w')$.
The more similar two words are, the larger penalty there would be for distribution difference. Clearly, we prefer the topic distribution with smaller $R(\beta)$.

In WR-LDA, we try to maximize function $O(\alpha,\beta)$ which is the combination of $L(\alpha,\beta)$ and $-R(\beta)$
instead of just optimizing log likelihood $L(\alpha,\beta)$\footnote{The details of $L(\alpha,\beta)$ is shown in Appendix A.} in standard LDA.
\begin{equation}
O(\alpha,\beta)=\lambda L(\alpha,\beta)-(1-\lambda)R(\beta)
\end{equation}
where $\lambda$ is the parameter that balances the likelihood and loss function. When $\lambda=1$, our model degenerates into standard LDA model. Similar idea can also been found in Zhang et al. (2010) for bilingual topic extraction and Mei et al.(2008) for topic network construction.
\subsection{Inference}
In this subsection, we describe the variational inference for WR-LDA. We emphasize on the part that is different from LDA and briefly describe the part that is similar.
We use the variational distribution $q(\theta,z|\gamma,\phi)$ to approximate posterior distribution $p(\theta,z|w,\alpha,\beta)$ and use variational EM algorithm for inference.

{\bf E-step}: The E-step of WR-LDA tries to find the optimizing values of the variational parameters $\{\gamma,\phi\}$.
Since the loss function $R(\beta)$ does not involve $\gamma$ and $\phi$,
E-step of WR-LDA is the same as that of standard LDA. We update $\{\gamma,\phi\}$ according to the following equations.
\begin{equation}
\phi_{nk}=\beta_{kw_n}exp\{E_q[\log(\theta_k)|\gamma]\}\\
\end{equation}\label{equ:update1}
\begin{equation}
\gamma_k=\alpha_k+\sum_n\phi_{nk}\\
\end{equation}\label{equ:update2}
where $E_q[\log(\theta_k)|\gamma]=\Psi(\gamma_k)-\Psi(\sum_k\gamma_k)$ and $\Psi(\cdot)$ is the first derivative of $\log\Gamma$ function. The E-step of WR-LDA is shown in Figure 2 (Blei et al., 2003).\par
\begin{figure}[!ht]
\centering
\begin{tabular}{p{7.5cm}}\\\hline\hline
1. initialize $\phi_{nk}^0:=1/k$ for all $k$ and $n$\\
2. initialize $\gamma_k:=\alpha_i+N/K$\\
3. {\bf repeat}:\\
~~~~for $n\in [1,N]$:\\
~~~~~~~~~for $k\in [1,K]$:\\
~~~~~~~~~~~~~~update $\phi_{nk}^{t+1}$ according to Equ.7\\
~~~~~~~~~normalize $\phi_n^{t+1}$\\
~~~~~~~~~update $\gamma^{t+1}$ according to Equ.8\\
~~~~{\bf until convergence}\\\hline\hline
\end{tabular}
\caption{E-step of WR-LDA}
\end{figure}
{\bf M-step}: We try to maximizes the resulting lower bound of $O(\alpha,\beta)$ shown in Equ.6 with respect to the model
parameters $\alpha$ and $\beta$. \par
{\bf Update $\beta$}:
By considering the Lagrange multipliers corresponding to the constraints that $\sum_w\beta_{kw}=1$ and $\sum_k \alpha^m_k=1$, we define $T(\alpha,\beta)$ as follows:
\begin{equation*}
\begin{aligned}
&T(\alpha,\beta)=\\
&O(\alpha,\beta)+\sum_kt_k(\sum_w\beta_{kw}-1)+\sum_dt_d(\sum_k\alpha^m_k-1)
\end{aligned}
\end{equation*}
So given $\{\gamma,\phi,\lambda\}$ we have
\begin{equation}
\frac{\partial O(\alpha,\beta)}{\partial \alpha_k}=0~~~~\frac{\partial T(\alpha,\beta)}{\partial \beta_{kw}}=0
\end{equation}

In LDA, $\beta$ is updated in the M-step with $\beta_{kw}\propto \sum_m\sum_{n\in m}\phi_{d_nk}{\bf 1}(w_{d_n}=w)$. However in WR-LDA,  because of the incorporation of loss function $R(\beta)$, the derivative of $T(\alpha,\beta)$ with respect to $\beta_{kw}$ depends on $\beta_{kw'}$, where $w'\neq w$, as shown in Equ.10.
\begin{equation}
\begin{aligned}
&\frac{\partial T(\alpha,\beta)}{\partial \beta_{kw}}=\lambda\frac{1}{\beta_{kw}}\sum_m\sum_{n\in m}\phi_{d_nk}{\bf 1}(w_{d_n}=w)\\
&~~~~~~~~~~~~~~-2(1-\lambda)\sum_{w'}\kappa_{w,w'}(\beta_{kw}-\beta_{kw'})+t_k
\end{aligned}
\end{equation}
To achieve the optimal value of $\beta$, we use Newton-Raphson method for optimization. $\beta_k$ is a $1\times V$ vector. The update for $\beta_k$ in Newton-Raphson is as follows:
\begin{equation}
\beta_k^{t+1}=\beta_k^{t}-H(\beta_k)^{-1}\bigtriangledown T(\beta_k)
\end{equation}
where $\bigtriangledown L(\beta_k)$ is the gradient of function $T(\alpha,\beta)$ with regard to $\beta_{kw_i}, i\in[1,V]$.
$H(\beta_k)$ is a Hessian matrix.
\begin{equation}
\begin{aligned}
&H(\beta_k)(i,j)=\frac{\partial^2 T(\alpha,\beta)}{\partial\beta_{kw_i}\partial\beta_{kw_j}}=-2(1-\lambda)\kappa_{w_i,w_j}\\
&-\frac{\lambda\delta(i,j)}{\beta^2_{kw_i}}\sum_m\sum_{n\in m}\phi_{nk}{\bf 1}(w_n=w_i)
\end{aligned}
\end{equation}
If the size of vocabulary V is $10^4$, $H(\beta_k)$ would be a $10^4\times 10^4$ matrix, the inverse of which would be too expensive to be calculated directly. To solve this problem, we adopt the strategy proposed in Mei et al.(2008)'s work, where we only need to find the value of $\beta^{n}$ that makes $T(\alpha^{n},\beta^{n}|\lambda^{n},\phi^{n},\gamma^{n})>T(\alpha^{n-1},\beta^{n-1}|\lambda^{n},\phi^{n},\gamma^{n})$
instead of getting the
local maximum of $T(\alpha^{n},\beta^{n}|\lambda^n,\phi^n,\gamma^n)$ at each M-step.
In each iteration of M-step, we firstly set the value of $\beta^{n,0}$ to the value which maximizes $L(\alpha,\beta)$, just as in Standard LDA.
\begin{equation}
\beta_{kw}^{n0}\propto \sum_m\sum_{n\in m}\phi_{d_nk}{\bf 1}(w_{d_n}=w)
\end{equation}
Then we iteratively obtain $\beta^{n,1}, \beta^{n,2},..., \beta^{n,t}$ according to Equ.14 until the value of $T(\alpha,\beta)$ drops.
\begin{equation}
\begin{aligned}
\beta_{kw}^{n,t}=\rho\beta_{kw}^{n,t-1}+(1-\rho)\frac{\sum_w'\kappa_{ww'}\beta_{kw'}^{n,t-1}}{\sum_w'\kappa_{ww'}}
\end{aligned}
\end{equation}
Clearly, $\sum_w\beta_{kw}=1$ and $\beta_{kw}\geq 0$ always hold in Equ.14.
Equ.14 can be interpreted as follows: when $\rho$ is set to 0, it means that the updating value of $\beta_{kw}$ is totally decided by its neighbors. In graph harmonic algorithm (zhu2003semi), Equ.14 is just the optimization of $Loss function~R(\beta)$ when $\rho$ is set to 0. So we try to optimize $O(\alpha,\beta)$ by firstly decreasing the value of $Loss function~R(\beta)$ until $O(\alpha,\beta)$ decreases.

{\bf update $\alpha$}: The first derivative of $T(\alpha,\beta)$ in regard with $\alpha_k$ also depends on $\alpha_{k'}, k\neq k'$, we also use Newton-Raphson method for optimization. Since the Hessian matrix $H(\alpha)$ is the form of:
\begin{equation*}
H(\alpha)=diag(h)+{\bf 1}z{\bf 1}^T
\end{equation*}
where $h_k=-M\Psi'(\alpha_k)$ and $z=M\Psi'(\sum_k\alpha_k)$the inverse of $H(\alpha)$ can be easily calculated (Blei et al., 2003):
\begin{equation*}
H^{-1}(\alpha)=diag(h)^{-1}-\frac{diag(h)^{-1}{\bf 11^T}diag(h)^{-1}}{z^{-1}+\sum_kh_k^{-1}}
\end{equation*}
$\alpha$ can be updated as follows:
\begin{equation}
\alpha^{t+1}=\alpha^{t}-H(\alpha)^{-1}\bigtriangledown T(\alpha)
\end{equation}
Multiplying by the gradient, we easily obtain the $i^{th}$ component of matrix $H(\alpha)^{-1}\bigtriangledown T(\alpha)$ (Blei et al., 2003).
\begin{equation}
[H(\alpha)^{-1}\bigtriangledown T(\alpha)]_i=\frac{[\bigtriangledown T(\alpha)]_i-c}{h_i}
\end{equation}
where $c=\frac{\sum_k[\bigtriangledown T(\alpha)]_k/h_k}{z^{-1}+\sum_kh_k^{-1}}$
\begin{figure}
\centering
\begin{tabular}{p{7.5cm}}\hline\hline
{\bf E-step}: For each document, find optimizing values of variational parameters $\{\gamma,\lambda,\phi\}$\\
{\bf M-step}: Maximize the resulting lower bound of $O(\alpha,\beta)$ with respect to the model
parameters $\{\alpha,\beta\}$ as follows:\\
~~~update $\beta$ based on Equ. 13 until convergence\\
~~~update $\alpha$ based on Equ. 14 until convergence\\\hline\hline
\end{tabular}
\caption{variational EM algorithm for WR-LDA model.}
\end{figure}
\section{Experiments}
In this section, we compare WR-LDA with standard LDA and other baselines in multiple applications.
\subsection{Text Modeling}
We firstly compare text modeling of the WR-LDA with standard LDA on the 20
Newsgroups data set\footnote{http://qwone.com/~jason/20Newsgroups/} with a standard list of 598 stop words removed. 

\begin{figure*}
\begin{minipage}[t]{0.5\textwidth}
\centering
\includegraphics[width=2.2in]{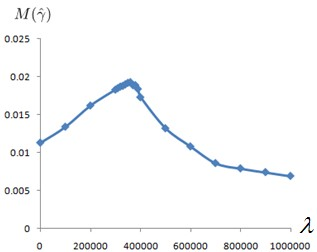}
\caption{Tuning $\lambda$(20News-Group data)}\label{fig:tune}
\end{minipage}%
\begin{minipage}[t]{0.5\textwidth}
\centering
\includegraphics[width=2.2in]{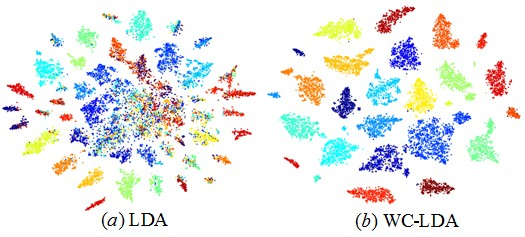}
\caption{t-SNE 2D embedding (a)LDA (b)WR-LDA}\label{fig:tsne}
\end{minipage}
\end{figure*}

We fit the dataset to a 110 topics. In WR-LDA, we need to tune $\lambda$. 
To evaluate the performances of WR-LDA with different $\lambda$, we have the following function.
\begin{equation}
\begin{split}
M(\hat{\gamma})=&\frac{1}{C_{1}}\sum\limits_{i=1}^{20}\sum_{d\in G_{i}}\sum_{d^{'}\in G_{i}}\zeta_{2}(KL(\hat{\gamma}^{d}||\hat{\gamma}^{d^{'}})) \\
& + \frac{1}{C_{2}}\sum\limits_{i=1}^{20}\sum_{j\ne i}\sum_{d\in G_{i}}\sum_{d^{'}\in G_{j}}\zeta_{1}(KL(\hat{\gamma}^{d}||\hat{\gamma}^{d^{'}}))
\end{split}
\end{equation}
$G_{i}$ denotes the collection of document with the label $i$; $\hat{\gamma}_{k}^{d}=\gamma_{k}^{d}/M_{d}$. $KL(p||q)$ represents the KL distance between two distributions. $\zeta_{1}(x)=e^{x}/(1+e^{x})$ and $\zeta_{2}(x)=1-\zeta_{1}(x)$. $C_{1}$ and $C_{2}$ are normalization factor. $\lambda$ varies from $0$ to $10^6$ with an interval of $10^5$. The results are shown in Fig \ref{fig:tune}. In the following experiments, we set $\lambda$ as $3.6*10^5$. 

\begin{figure*}
\centering
\includegraphics[width=0.9\textwidth]{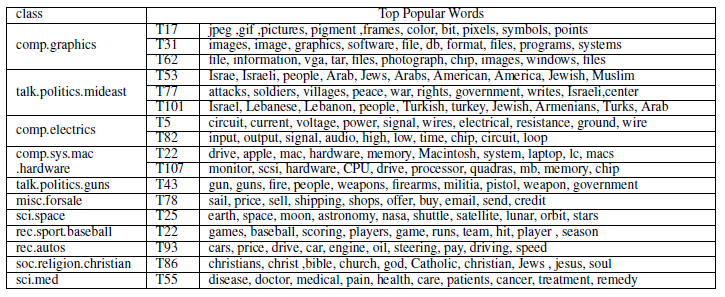}
\caption{Top words in selected classes from 20NewsGroup by WR-LDA}\label{fig:top}
\end{figure*}

Fig \ref{fig:tsne} shows the 2D embedding of the expected topic proportions of WR-LDA($\lambda=3.6*10^5$) and LDA using the t-SNE stochastic neighborhood embedding (van der Maaten and Hinton, 2008).
Fig \ref{fig:top} shows some of the results.

\subsection{Regression}

\begin{figure}
  \begin{center}
    \includegraphics[width=0.48\textwidth]{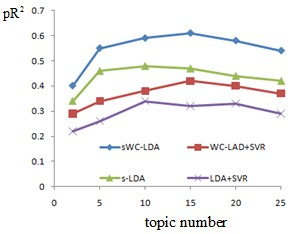}
  \end{center}
\caption{$pR^2 prediction$}\label{fig:reg}
\end{figure}

We evaluate supervised version of WR-LDA on
the hotel reviews dataset from TripAdvisor\footnote{http://www.tripadvisor.com/} . As in Zhu and Xing (2010), we take logs of the response
values to make them approximately normal. Each
review is associated with an overall rating score and
five aspect rating scores for the aspects: Value,
Rooms, Location, Cleanliness, and Service.
In this paper, we focus on predicting the overall
rating scores for reviews.

We compare the results from the following
approaches: LDA+SVR, sLDA, WR-LDA+SVR
and sWR-LDA. For LDA+SVR and WC-
LDA+SVR, we use their low dimensional
representation of documents as input features
to a linear SVR and denote this method. The
evaluation criterion is predictive $R^2$ (pR2) as
defined in the work of Blei and McAuliffe (2007).

Figure \ref{fig:reg} shows the predictive $R^2$ scores of different models. First, since supervised topic models 
can leverage the side information (e.g., rating scores) to discover latent topic representations, they
generally outperform the decoupled two-step procedure as adopted in unsupervised topic models. Secondly, we can see that sWR-LDA outperforms sLDA and WR-LDA+SVR outperforms LDA+SVR.
 
\begin{figure*}
\centering
\includegraphics[width=0.96\textwidth]{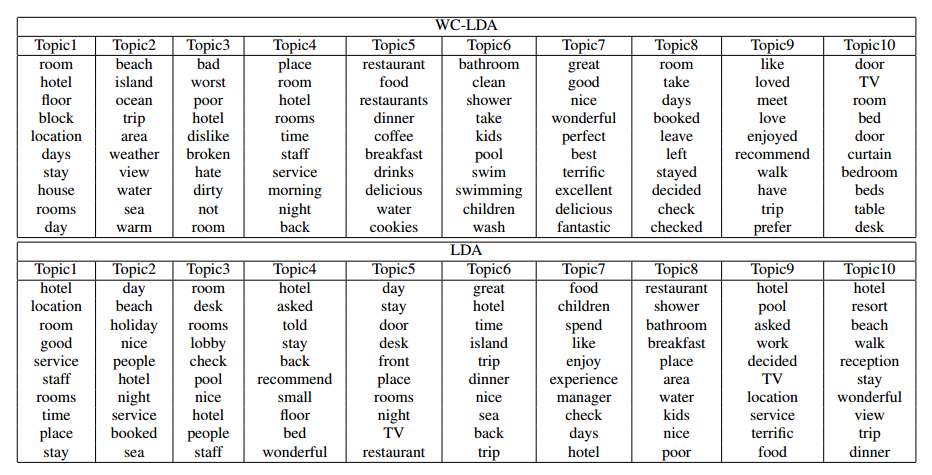}
\caption{WR-LDA \& LDA on TripAdvisor dataset}\label{fig:tab2}
\end{figure*}

 From the top words of different topics shown at Fig\ref{fig:tab2}, we can easily explain why sWR-LDA is better than sLDA. For sWR-LDA, part
of topics show a regular positiveness/negativeness
pattern. Topic 3 is on the negative aspects of a
hotel while Topic 7 and Topic 9 describe positive aspects. This is because word correlation is
considered in WR-LDA. For example, positive vocabularies such as great, good and wonderful,
they have similar semantic meanings and thus high
edge weight. So they are very likely to be generated from the same topic. The topics discovered by
LDA do not show a regular pattern on positiveness
or negativeness, resulting in low predictive $R^2$.
Moreover, we can see that topics are more coherent in WR-LDA. Topic 2 talks about sea sceneries,
Topic 5 talks about food, Topic 6 talks about conditions of bathroom or swimming pool and Topic
10 talks about room objects. However, topics in
LDA do not show such clear pattern.

\section{Cross-Lingual Topic Extraction}
\subsection{Dataset Construction}
To qualitatively compare our approach with the baseline
method, we test different models on Cross-Lingual Topic Extraction task in different respects.
The data set we used in this
experiment is selected from a Chinese news website Sina(新浪)\footnote{http://www.sina.com.cn}
and Google News\footnote{https://news.google.com}. Selected news talks about 5 topics categorized by news websites such as "Sports(体育)", "Entertainments(娱乐)", "World(国际)", "Science(科学)" and "Business(财经)".
Each Chinese news passage is associated with an English news passage which talks about the same event (i.e. Passage entitled "土耳其民众对政府改造广场抗议蔓延至全国" vs Passage entitled "Protests no Turkish Spring, says PM Erdogan").
We select 1000 English
passages and 1000 Chinese passages during the period from March. 8th, 2001 to Jun. 23th, 2013, including \{Sports:360, WordNews:296, Bussiness:92, Science:68, and Entertainments:184\}.

\subsection{Details}
Let $C_c$ denote Chinese news collection, where $C_c=\{d_c^1,d_c^2,...,d_c^N\}$, $N=1000$. $C_e$ is the English news collection $C_e=\{d_e^1,d_e^2,...,d_e^N\}$, and $d_c^i$ is the correspondent passage of $d_e^i$, $i\in [1,N]$. 
We firstly removed infrequent words\footnote {Words that occurred less than 3 times in the corpus.} and frequent words \footnote {Words that occurred more than M/10 times in the corpus, where M is the total number of documents.}. A bilingual Chinese-English dictionary
would give us a many-to-many mapping between the vocabularies of the two languages.  If one word can be potentially translated into another word, the two words
would be connected with an edge. Specifically, let $w_e$ denote an English word and $w_c$ denotes a Chinese word. $C_{w_e}$ denotes the set of Chinese words that are translated from $w_e$, $C_{w_e}=\{w_c|e(w_c,w_e)\neq0\}$
\begin{equation}
\kappa(w_e,w_c)=\left\{
\begin{aligned}
&1~~if~~w_c\in C_{w_e}\\
&0~~if~~w_c\not\in C_{w_e}
\end{aligned}
\right.
\end{equation}

\subsection{Evaluation}
Part of the results are presented in Table \ref{tab:re}.
Since commonly used measures for LDA topic model
such as perplexity can not capture whether topics are coherent or
not, we use the measures in Petterson el al. (2011)'s work. 

\begin{table*}
\centering
\footnotesize
\begin{tabular}{rccl}\hline
Topic 1&Topic 5&Topic 6&Topic 11\\\hline
James&Korea&习近平(Xi Jinping)&巴塞罗那(Barcelona) \\
詹姆斯(James)&North&China&Champions\\
科比(Kobe)&朝鲜（North Korea&government&Ferguson\\
Kobe&Kim&Jin Ping&Madrid\\
Heat&Jong-un&Xi&拜仁(Bayern)\\
rebound&金正恩(Kim Jong-un)&李克强(Li Keqiang)&半决赛(semi final)\\
热火(Heat)&中国(China)&prime&retire\\
Lakers&United&China&Barcelona\\
game&韩国(South Korea)&选举(selection)&score\\
basketball&talk&Chinese&欧冠(UEFA Champions)\\
湖人(Lakers)&South&chairman&小罗(Ronaldo)\\
Durant&Obama&主席(chairman)&曼联(Manchester United)\\
杜兰特(Durant)&missile&代表(representative)&goal\\
score&导弹(missile)&Keqiang&英超(English Premier League)\\
hurt&United&人民(People)&多特蒙德(Dortnumd)\\
Heat&States&马克思(Marx)&German\\
Thunder&defend&政治局(political bureau)&Messi\\
雷霆(Thunder)&军事(military)&国家(country)&guard\\
season&外交部(Foreign Ministry)&selection&Ronaldo\\
洛杉矶(Los Angeles)&military&people&进球(score)\\\hline
\end{tabular}
\centering
\caption{Top Word in selected topic extracted from WR-LDA in Cross-Lingual task}\label{tab:re}
\end{table*}

We compare the topic distributions of each Chinese news passage with its
correspondent English pair based on the following measures:\\
\begin{itemize}
\item Mean $l_2$ Distance (L2-D):\\$\frac{1}{N}
\sum_{i=1}^{N}((\sum_{k=1}^{N}\hat{\gamma}_k^{d_c^i}-\hat{\gamma}_k^{d_e^i})^2)^{1/2}$
\item Mean Hellinger Distance (H-D):\\
$\frac{1}{N}
\sum_{i=1}^{N}\sum_{k=1}^{N}((\hat{\gamma}_k^{d_c^i})^{1/2}-(\hat{\gamma}_k^{d_e^i})^{1/2})^2$
\item Agreements on first topic: (A-1)\\
$\frac{1}{N} \sum_{i=1}^{N}I(argmax_k\hat{\gamma}_k^{d_c^i}=argmax_k\hat{\gamma}_k^{d_e^i})$
\item Mean number of agreements in top 5 topics (A-
5):\\
$\frac{1}{N} \sum_{i=1}^{N}argeement(d_c^i,d_e^i)$,
where agreements $argeement(d_c^i,d_e^i)$ is the cardinality
of the intersection of the 5 most likely topics
of $d_c^i,d_e^i$.
\end{itemize}

Clearly, we prefer smaller values
of L2-D and H-D and larger values of A-1 and A-
5. We compare the performances from following
approaches:\\
\begin{itemize}
\item WR-LDA1: Only Chinese-English correlations
are considered and the weights of edges
between English-English words, Chinese-Chinese words are 0.
\item  WR-LDA2: All word correlations are considered.
\item DC model: Approach proposed by Petterson
et al.(2011), where word correlations are incorporated
into a Beta prior using a logistic smooth
function.
\item LDA: the standard LDA topic model.
\end{itemize}

\begin{figure}
  \begin{center}
\includegraphics[width=0.48\textwidth]{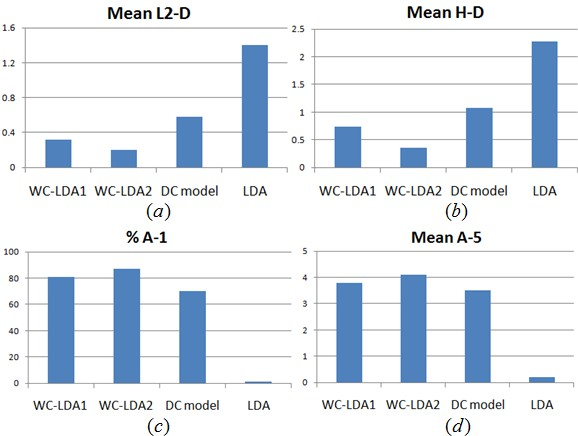}
  \end{center}
\caption{Comparison of topic distributions }\label{fig:comp}
\end{figure}

From Figure~\ref{fig:comp}, LDA achieves the
worst results as expected because it can hardly
detect the topics shared by bilingual documents
due to the reason that vocabularies from different
languages hardly co-appear in the same passages.
WR-LDA1 is better than the DC model, which considers
word correlations only in the prior
construction, rather than in the algorithm. WR-LDA2,
which consider word correlations fully,
achieves better results than WR-LDA1,
which partly considers word correlations.

\begin{figure*}
\begin{minipage}{0.9\textwidth}
\centering
\includegraphics[width=4.8in]{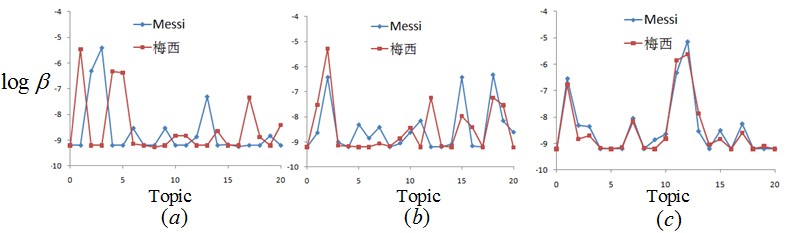}
\caption{Topic-Word distribution for Messi-梅西 (a)LDA (b)DC (c) WR-LDA}
\end{minipage}%
\\
\begin{minipage}{0.9\textwidth}
\centering
\includegraphics[width=4.8in]{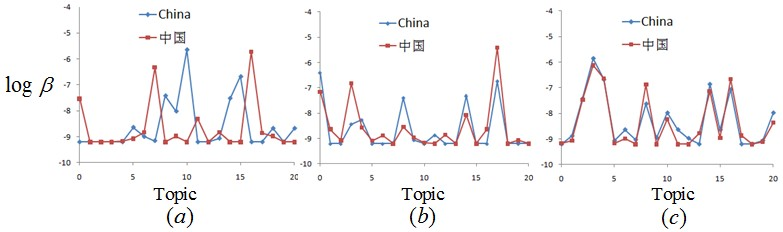}
\caption{Topic-Word distribution for China-中国 (a)LDA (b)DC (c) WR-LDA}
\end{minipage}
\end{figure*}

For further illustration, we randomly choose two
pairs of words, ”Messi” and ”梅西” (Messi in
Chinese), which is the name of a famous Argentine
soccer player, and ”China” and ”中国(China
in Chinese)”. Since each pair of words have the
same meaning, we prefer that they have the similar
probabilities generated from the same topic. Figure
8 shows the probability that ”Messi” and ”梅西” 
generated from different topics and Figure 9
shows ”China” and ” 中国”. The x-axis is the index
to the topic and y-axis is the log-likelihood of
probability. We can see that in LDA, words within
each pair are always negative correlated. This can
be easily explained by the fact that Chinese words
and English words never co-appear in a document.
So if a Chinese word has large probability generated
by one topic, it means this top is a Chinese
dominated topic and most English words would be
excluded. Word pairs in DC model are correlated
but not as strong as that in WR-LDA. Since DC
model models word correlation only in the prior,
such influence can be diluted while Gibbs sampling
goes on. In WR-LDA, word correlations are
considered in the algorithm and during each iteration,
algorithm will fix the gap between correlated
words. That is why WR-LDA outperforms DC
model.

\section{Conclusion}

In this paper, we present WR-LDA, a revised
version of LDA topic model that incorporates
word correlations. Experiments on text modeling,
review rating prediction and cross-lingual topic
modeling demonstrate the effectiveness of our
model.
There are two disadvantages of WR-LDA when
compared with LDA. (1) The value of parameter
lambda involved in the model is hard to tune while
LDA there is no additional parameter involved.
(2) Due to the introduction of Penalty function in
WR-LDA, we have to keep record of parameter
$\phi$ in varational inference for all documents, which
largely increase the cost of both memory and time

 \end{CJK*}
\bibliographystyle{abbrv}
\bibliography{mybiblio}

\end{document}